\pdfoutput=1
\documentclass[11pt]{article}

\usepackage[]{acl}

\usepackage{times}
\usepackage{latexsym}
\usepackage{amssymb}

\usepackage[T1]{fontenc}
\usepackage[utf8]{inputenc}
\usepackage{fullpage}
\usepackage[T1]{fontenc} 
\usepackage{epsf} 
\usepackage{graphics} 
\usepackage{psfrag,xspace}
\usepackage{color,etoolbox}
\usepackage[ruled,vlined]{algorithm2e}
\usepackage{mathtools}  
\usepackage{thm-restate}
\usepackage{dsfont}
\usepackage{booktabs} 
\usepackage{microtype}

\usepackage{inconsolata}
\usepackage{comment}

\usepackage{xcolor}
\definecolor{oceanblue}{rgb}{0.4, 0.7, 0.85} % A deep blue with a hint of green
\definecolor{sunsetorange}{rgb}{0.95, 0.65, 0.4} % A vibrant orange with a touch of red
\definecolor{mintgreen}{rgb}{0.8, 1.0, 0.8} % A light, refreshing green
\definecolor{lavenderpurple}{rgb}{0.85, 0.75, 1.0} % A soft purple with a hint of blue

\usepackage{colortbl}
\newcommand{\orchestrallm}{\textsc{OrchestraLLM}\xspace}
\newcommand{\correctionlm}{\textsc{CorrectionLM}\xspace}

\newcommand{\SLM}{Llama-3-8B\xspace}

\newcommand{\LLM}{GPT-4o\xspace}
\input{Definitions}
\title{CorrectionLM: Self-Corrections with
SLM for Dialogue State Tracking}

 \author{Chia-Hsuan Lee \\
 University of Washington\\
   \texttt{chiahlee@uw.edu} \\\And
   Hao Cheng \\
   Microsoft Research \\
   \texttt{chehao@microsoft.com}\\\And
   Mari Ostendorf\\
   University of Washington\\
   \texttt{ostendor@uw.edu
}}

\begin{document}
\maketitle
\begin{abstract}
Large language models (LLMs) have demonstrated self-improvement capabilities via feedback and refinement, but current small language models (SLMs) have had limited success in this area. Existing correction approaches often rely on distilling knowledge from LLMs, which imposes significant computation demands. In this work, we introduce \textbf{\correctionlm}, a novel correction framework that enables SLMs to self-correct using in-context exemplars without LLM involvement. Applied to two dialogue state tracking (DST) tasks in low-resource settings, \correctionlm achieves 
results similar to a state-of-the-art LLM at a small fraction of the computation costs.

\end{abstract}

\section{Introduction}

% As noted in Chapter~\ref{ch:background}, what constitutes an SLM has been increasing in size due to technology improvements, with increasing ability to do in-context learning. In this chapter, we take advantage of this with Llama3~\citep{dubey2024llama}, which with 8B params can be considered small relative to other frontier models such as GPT-4 and GPT-3.5.
Recently, large language models (LLMs) have demonstrated strong reasoning abilities by providing feedback on their own outputs and subsequently refining them based on that feedback~\citep{shinn2024reflexion,madaan2024self,huang2023large,saunders2022self}. This is especially true for tasks requiring multi-step reasoning (code and math reasoning tasks). However, these capabilities of generating feedback and refinement are less commonly observed in small language models (SLMs)~\citep{saunders2022self,selfee2023}.

To enable the abilities of self-critique and self-refinement of SLMs, previous research has focused on distilling knowledge from LLMs. Typically, this involves fine-tuning SLMs on improvement demonstrations generated by LLMs~\citep{shridhar2023distilling,selfee2023} and has proven to improve the self-improvement abilities of SLM.

However, ~\citet{yu-etal-2024-teaching} observe that naively training SLMs on LLM improvement demonstrations can hurt task performance, since SLMs may have different error modes, and learning from LLM mistakes may be less beneficial. To address this, they propose generating reasoning trajectories with SLMs and using LLMs to provide feedback and refinement before fine-tuning the SLMs on these edited trajectories. However, this approach still heavily relies on LLM involvement.
% \hao{It's hard to say whether using human-labeled data is better than using feedback from LLMs. Both are expensive.}\michael{I agree. But deploying LLM requires lots of computations which might not be feasible for all developers.     }
% \michael{But related, do we want to highlight the contrast of us not having to use reasoning trajectories and feedback/reflection? The reason I didn't write that down is if we say we didn't use those because DST is easier than math/code, that will make us sound weak.}
% \hao{Definitely no. most DST predictions can be done in one-shot but math/code ones are more complex where long trajectories are necessary to guide the model for better predictions. }
In this work, we introduce a novel self-improvement framework, \correctionlm, that finetunes an SLM and makes corrections using in-context exemplars, without involving any LLMs. We demonstrate the effectiveness of \correctionlm on dialogue state tracking (DST), a task that extracts user intents from multi-turn conversations.
% As discussed in Chapter~\ref{ch:background}, the definition of an SLM has evolved due to technological advancements, with even larger SLMs now capable of in-context learning. In this chapter, we leverage this progress by utilizing Llama3, which, with 8 billion parameters, is considered small relative to frontier models such as GPT-4.

In this work, we target low-resource settings and use only 5\% of the training set for experiments. We first randomly sample a few examples as in-context learning prompts for the SLM to generate dialogue state predictions for the remaining data. This step is intended to capture the errors that SLMs are prone to during in-context learning inference. We then finetune the SLM in a parameter-efficient way on the conversations along with self-generated predictions and gold labels, resulting in an SLM capable of making corrections. Unlike prior work, our method does not rely on LLMs to provide feedback or generate predictions. Additionally, our method does not require an external verifier (such as execution environment~\citep{chencodet} or heuristic filter~\citep{shinn2024reflexion}), as the model internally decides whether to make changes to the initial predictions. Experimental results on two benchmarks in low-resource settings show that \correctionlm can effectively make corrections leveraging in-context exemplars and achieves superior computational efficiency compared to the previous state-of-the-art. 

\section{Low Resource Dialogue State Tracking}
A task-oriented dialogue (TOD) is composed of a series of exchanges between two parties, where each exchange begins with a user input and is followed by a system response. We refer to each exchange as a turn, forming a sequence \( U_1, A_1, \dots, U_T, A_T \), with \( U_t \) representing the user's utterance and \( A_t \) the system's reply at turn \( t \). During the \( t \)-th turn, the user offers a new utterance \( U_t \), to which the system responds with \( A_t \). The dialogue context at turn \( t \) is defined as \( C_t = \{ U_1, A_1, \dots, A_{t-1}, U_t \} \). The primary objective of DST is to extract and convert task-relevant details from these user-system interactions into structured representations, known as dialogue states, ensuring that the system can effectively address the user’s needs. More concretely, given context $C_t$, the goal is to predict the dialog state at time $t$, a collection of slot-value tuples $\{(s_i^t, v_i^t); i=1,\ldots , n\}$, where $n$ is the number of possible slots.  Using the language modeling approach, the dialogue state is predicted as:
\begin{equation}
    DST_t = LM(T, DST_{t-1}, \tilde{C}_t),
\end{equation}
where $T$ is the schema table for target domains and $\tilde{C}_t$ is typically a truncated version of the context.
Specifically, our implementation predicts the turn level belief $TLB_t$ (the set of new or updated slot-value pairs), assuming initial values of each slot is null, and then aggregating $DST_{t-1}$ with $TLB_t$ to get $DST_t$.

This work focuses on a low-resource setting where only a limited amount of training data is available, in which case in-context learning (ICL) is useful. Following IC-DST \citep{hu2022context}, the turn-level belief is predicted as:
\begin{equation}
     TLB_t = \text{LM} (T, E_{1:k}, DST_{t-1}, \tilde{C}_t)
     \label{eqn:TLB_base}
\end{equation}
where $E$ is a sequence of $k$ examples of inputs and gold TLB outputs 
\begin{equation}
    E_i = (DST_{\tau_i-1}, \tilde{C}_{\tau_i} , TLB^*_{\tau_i})
\end{equation}
that are retrieved from the training data based on similarity of embeddings of the input context. Here, the notation $\tau_i$ is used to be clear that the time of the user utterance in example $i$ may be different from $t$. In \citet{hu2022context}, the most effective retriever is an embedding model initialized with SenBERT~\cite{reimers-2019-sentence-bert}, and subsequently fine-tuned to maximize similarity of the gold TLBs based on the F-score. In our study, we adopt this procedure to develop our in-context example retriever.

\section{Correction Approach with SLMs}

Our proposed \correctionlm leverages in-context learning in a two-pass generative LM approach.
As shown in \autoref{fig:correctionlm-inf}, it incorporates two SLMs: the inference SLM and the correction SLM.
For efficiency purpose, we use an off-the-shelf pretrained LM (a \textit{frozen} base LM) for the initial inference SLM and 
develop the second-pass correction SLM on top of the base LM using a parameter-efficient method fine-tuned for ICL-based correction. 
The limited training data is used in three ways: i) to provide examples for in-context learning in inference, ii) to tune the retriever for selecting examples, and iii) for fine-tuning the second-pass correction SLM.
In the following, we first describe the two pass inference in \S\ref{ssec:inference} and then provide details on how to train the correction SLM in \S\ref{ssec:training}.

\subsection{Two-Pass Inference}
\label{ssec:inference}

\begin{figure*}[t]
\centering
   \includegraphics[width=0.85\textwidth]{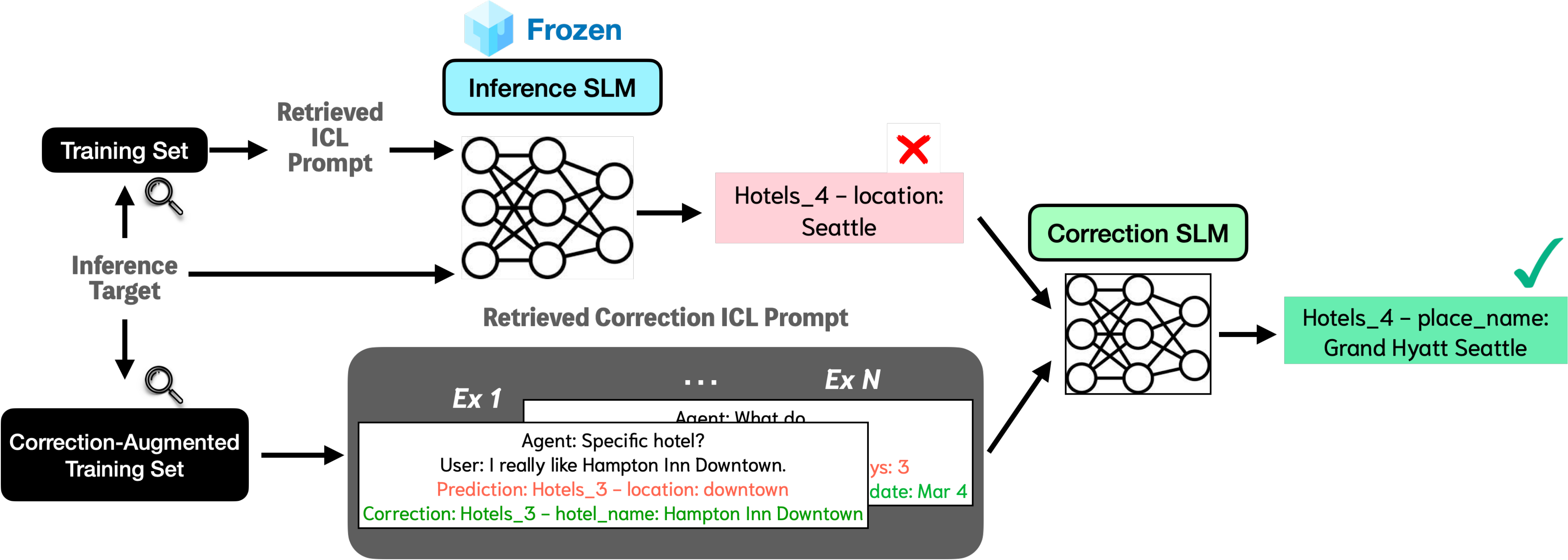}
   \caption{
 Illustration of the two-pass approach of \correctionlm. The first pass involves using the baseline ICL model, which prompts the SLM to predict TLB with retrieved in-context examples of inputs and gold outputs. The second pass prompts a second LM with examples of corrections.} 
   \label{fig:correctionlm-inf}
\end{figure*}

The two-pass DST prediction process of our method is depicted in \autoref{fig:correctionlm-inf}.
During the first pass, an initial TLB is generated for the current turn using the baseline IC-DST model as described in \autoref{eqn:TLB_base} built on top of the inference LM. % , with \text{infer\_SLM} as the LM. 
This step is done in the same fashion as \citet{hu2022context} with $k$ demonstration examples retrieved from the training set.
In the second pass, we design a new correction ICL for steering the model to make necessary corrections.
Concretely, we construct a correction prompt that includes the same $k$ examples from the previous pass but now augmented with correction demonstrations (including model initial prediction and correction pairs), and initial hypotheses (the initial TLB predicitions for the current input from the inference SLM). We then prompt the correction-tuned SLM (\text{correction SLM}) to refine the initial predictions made in the first pass.
Prompting details are included in Section~\ref{sec:expts}.
Unlike previous work \citep{yu2024teaching}, no rationale is used in our second pass prompting.

\subsection{Training}
\label{ssec:training}

\begin{figure*}[t]
\centering
   \includegraphics[width=0.85\textwidth]{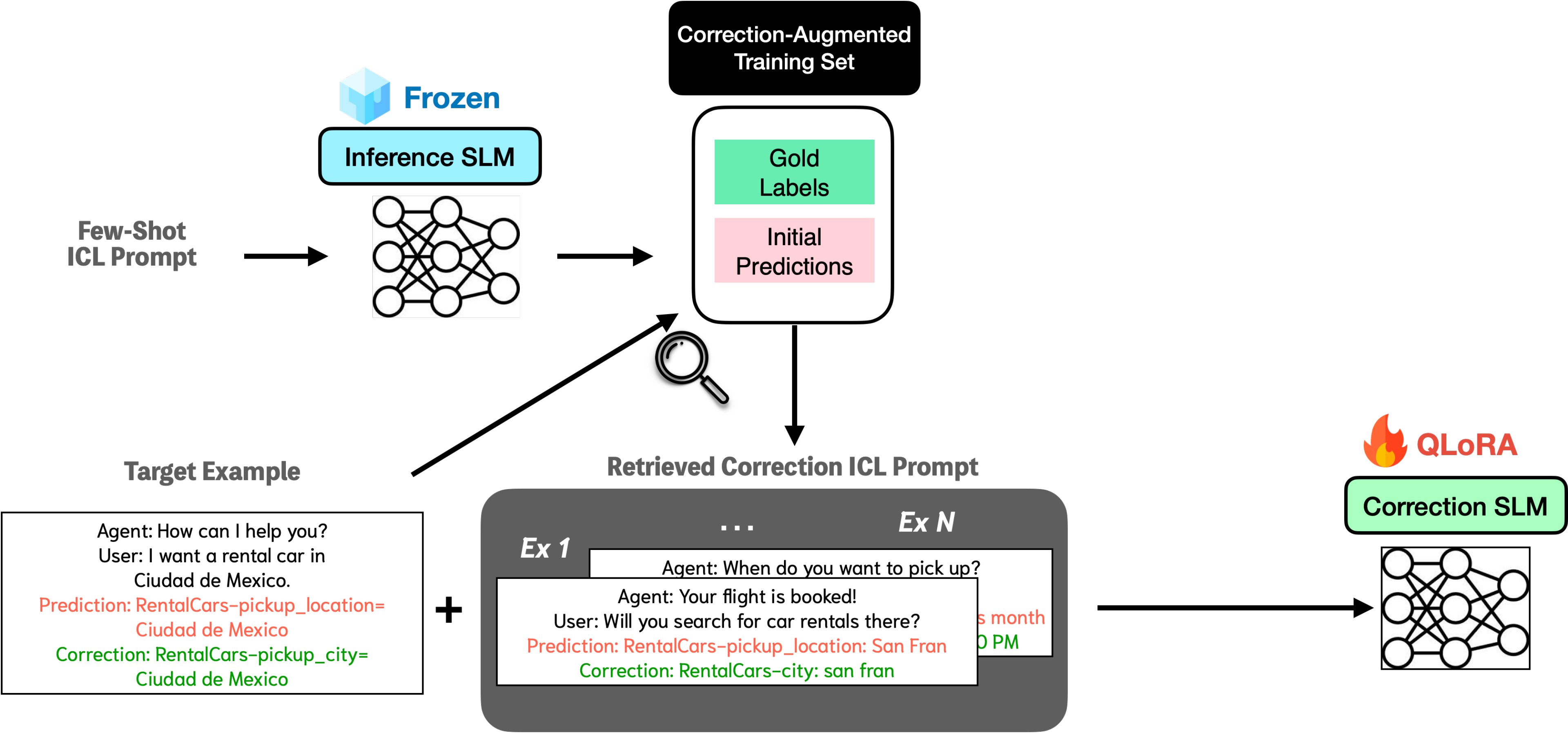}
   \caption{
Illustration of the \correctionlm training process. The first step is to prompt the first-pass inference SLM with a few in-context exemplars to produce predictions for each example in the training set. The second step is to finetune an SLM to generate the gold label given correction-augmented in-context exemplars and the target input and initial target prediction from the first step.} 
   \label{fig:correctionlm-train}
\end{figure*}

Note that both the inference SLM (built upon a frozen base LM using ICL) and the in-context demonstration retriever are developed in the same manner as outlined in \citet{hu2022context}.
Thus, we mainly focus on describing the training procedure for the correction SLM in this section.
Specifically, we develop a correction-augmented ICL schema for this purpose.

\paragraph{\bf Collecting Correction Demonstration}
The first step of the training is to obtain SLM predictions using ICL in order to provide supervision signals for the correction training. We randomly sample 5 fixed examples as task demonstrations and prompt \SLM to obtain predictions for every example in the training set. The prompts used in this step are consistent with those in the first step of the inference stage, except for the number and choice of the few-shot examples.

\paragraph{\bf Finetuning SLM to Learn to Self-Correct via In-Context Learning (Correction Tuning)}
Although we can directly finetune an SLM to correct errors in initial predictions, this strategy generally require a significant amount of training data, making it less effective in our low-resource setting, as we will demonstrate in \S\ref{sec:expts}.
Recent studies~\cite{min2022metaicl,chen2022meta} have shown that finetuning language models with in-context examples can improve their in-context learning performance when testing with few-shot examples. 
Building on these findings, we introduce a correction-augmented ICL tuning approach. 
Unlike traditional ICL methods that only consider the input and gold output, we also incoporate the model's (erroneous) self predictions into the demonstrations.
The SLM is then finetuned with a standard cross-entropy loss on the entire sequence of in-context correction examples.
To improve efficiency, we propose finetuning the model to correct its predictions using the parameter efficient QLoRA method~\citep{dettmers2024qlora}, where the parameters of the SLM are quantized before training and only the additional light-weight adapters are updated during the training process. 

%%MO: I moved this paragraph to experiments because I think it is only true for SGD.  You can put it back here if not
%In our initial experiments, we observed that finetuning the SLM was especially useful when correction targets involve domains not present in the training data, which motivated the use of domain-specific schema tables in the examplars (eqn.~\ref{eqn:examples-2b}) that are used in the prompt. 
%This approach seems to help the model learn to generalize and correct across domains. 

\section{Experiments}
\label{sec:expts}

\begin{table}[t]
    \small
    \centering
    \begin{tabular}{l@{\hskip3pt}c@{\hskip3pt}c}
    \toprule
        Dataset &  MultiWOZ &  SGD  \\
        \midrule
        \# Domains & 8 & 41  \\
        \# Dialogues & 8438 & 16142 \\
        \# Total Turns & 113556 & 329964   \\
         Avg. Turns per Dial. & 13.46 & 20.44  \\   
         Avg. Toks per Turn & 13.13 & 9.75   \\   
        \# Slots & 24 & 214 \\ 
        \# Slot Values & 4510 & 14139 \\ 
        \bottomrule
    \end{tabular}
    \caption{Experiment data summary. The numbers are computed on training splits of the datasets.
    \label{tab:data_sum}}
\end{table}
\subsection{Datasets and Evaluation}
We consider two DST datasets and a low-resource setting in which we use 5\% training set as the training data. Detailed statistics are presented in Table~\ref{tab:data_sum}. For both datasets, the low resource setting is defined as 5\% of the training data, consistent with prior reported work. 

\textbf{MultiWOZ} \cite{budzianowski2018multiwoz} is a comprehensive multi-domain task-oriented dialogue dataset comprising over 10,000 human-human written dialogues across 8 distinct domains. Since its initial release, several efforts have been made to enhance the quality of its annotations, leading to updated versions such as MultiWOZ 2.1 \cite{eric2020multiwoz} and MultiWOZ 2.4 \cite{ye2021multiwoz}. In our experiments, we utilize the latest version, MultiWOZ 2.4. 

\textbf{SGD \cite{rastogi2020towards}} 
is a task-oriented dialogue dataset consisting of over 16,000 multi-domain conversations across 41 domains, designed to evaluate out-of-domain performance. Notably, 15 out of the 21 domains in the test set are absent from the training set, and 77\% of dialogue turns in the test set include at least one domain not seen during training.
% \footnote{Due to the limited API budget, we couldn't experiment with LLM (e.g. GPT-4o) on a larger testing set}.

\textbf{Evaluation}
We report both turn-level scores (TLB~\cite{dey2022towards,hu2022context}) and dialogue-level scores (DST~\cite{henderson2014second}). DST metrics evaluate the accuracy of the cumulative dialogue states at each turn, while TLB metrics focus on local predictions at each turn independently, ignoring slot values from previous turns to factor out the effect of error propagation from previous turns.
%provide a more granular view of performance.

For both types of metrics, we use Joint Goal Accuracy (JGA) as established in prior work~\cite{wu2019transferable}. JGA assigns a score of 1 only when all slot values are correct. Thus, we also report F1 scores to capture partial correctness by matching subsets of slot values.

% \mari{I rewrote because: 1) I think you need to justify by consistency and not say it underestimates performance, and 2) I didn't understand what you wrote. Please edit if this is not correct.}
For the SGD dataset, we modified the TLB score to be consistent with the use of synonyms in JGA evaluation.
Specifically, we say the hypothesized slot-value pair is correct if the value corresponds to any of the specified synonyms for the gold value.
%find that the previous TLB evaluation scheme tends to underestimate system performance. This is due to the fact that SGD includes synonyms for the gold values as turn-level labels at various turns, while some systems may omit these values, treating them as redundant or repetitive. Therefore, we give the systems turn-level scores if their captured accumulative dialogue states include one of the synonyms of the gold values for the same slot.

\subsection{Experimental Setting}
\label{sec:implementation}
We use \SLM\footnote{https://huggingface.co/meta-llama/Meta-Llama-3-8B-Instruct} as the backbone SLM. We also use \LLM as the LLM for comparisons. We develop IC-DST 2.0 based on ~\citep{hu2022context}, but using \SLM or \LLM as the backbone language models instead of Codex.
In addition, there were some prompt configuration details specific to each dataset, because SGD has a much more complex schema and more turns per dialogue than MultiWOZ. 
%%MO: I think the memory constraint goes without saying, and I think that the larger amount of SGD data may have impacted the choices as well
%\footnote{This is due to the memory constraint during training Llama as SGD has more turns and schema tables.}. 

For MultiWOZ, we use $k$=10 examples in ICL prompts. The dialogue context $\tilde{C}_t$ includes only one agent-user turn pair.
The correct\_LM prompting structure is as follows:
\begin{equation}
     TLB_t = \text{correct\_SLM} (T, G_{1:k}, DST_{t-1}, \tilde{C}_t, H_t),
     \label{eqn:TLB_correct-mwoz}
\end{equation}
where $H_t$ is the first pass TLB hypothesis, and
\begin{equation}
    G_i = (DST_{\tau_i-1}, \tilde{C}_{\tau_i} , H_{\tau_i}, TLB^*_{\tau_i}).
    \label{eqn:examples-2a}
\end{equation}

For SGD, we use $k$=3 examples in ICL prompts. The dialogue context $\tilde{C}_t$ includes three agent-user turn pairs because of the more complex dialogues. 
Because of the large number of domains, a domain-specific schema table is added to the example $G'_i$, and the prompt order is changed accordingly:
\begin{equation}
     TLB_t = \text{correct\_SLM} (G'_{1:k}, T, DST_{t-1}, \tilde{C}_t, H_t),
     \label{eqn:TLB_correct-sgd}
\end{equation}
\begin{equation}
    G'_i = (T_i, DST_{\tau_i-1}, \tilde{C}_{\tau_i} , H_{\tau_i}, TLB^*_{\tau_i}).
    \label{eqn:examples-2b}
\end{equation}
This approach seems to help the finetuned model learn to generalize and correct across domains. 

For both datasets, following ~\citep{hu2022context}, the retriever is initialized with SenBERT (all-mpnet-base-v2) and finetuned with the 5\% training set.

\subsection{Baselines and Ablations}

We compare our \correctionlm performance to a single-pass ICL baseline, as well as to contrasting models that omit either the finetuning or the in-context examples.

\subsubsection{Single-Pass ICL}
We compare to a baseline that uses the same first pass SLM inference model without any corrections. We also report the results for the same approach with an LLM.

% \subsubsection{Zero-Shot Correction Baseline}
% The first correction baseline is prompting GPT-4o with only task instruction and schema table and no in-context exemplars.\footnote{We also experiment with Llama3 8B in this setting but it fails to produce meaningful outputs consistently.}

%%%%%%TODO: baseline
\subsubsection{In-Context Correction (Non-Finetuned)}
To show the benefit of finetuning, we report results for a system that prompts the same non-finetuned SLM with the prompts that include corrections, as specified in sec~\ref{sec:implementation},
%from equation~\ref{eq:fix_icl_mwoz}, equation~\ref{eq:fix_icl_sgd}, equation~\ref{eq:second_pass_mwoz} and equation~\ref{eq:second_pass_sgd}. 
except that the prompt format for the \correctionlm by includes a task description to instruct the LM to make corrections.
Again, we include a contrasting LLM version.

\subsubsection{Correction (Finetuned)}
To demonstrate the effectiveness of correction through in-context examples, we report results for a system that uses a finetuned SLM with prompts that 
%from equation~\ref{eq:finetune_target_mwoz} and equation~\ref{eq:finetune_target_sgd} excluding 
exclude the in-context examples $G_\mathit{1:k}$. 
%%MO: The sentence below seems redundant to me
%After finetuning, the model is then prompted with the first-pass and second-pass prompts during the inference stage, again excluding the in-context examples.

\subsubsection{\correctionlm}
We report results for two versions of \correctionlm. We finetune SLM using SLM predictions. We then make in-context corrections on SLM and LLM inference separately.  

% \begin{table*}[t]
%     \small
%     \centering
%     \begin{tabular}{l|l|l|cc}
%     \toprule
%         \textbf{Inference Model on Train}  & \textbf{Inference Model on Test}  & \textbf{Correction Model}  & \textbf{DST JGA/F1} & \textbf{TLB JGA/F1} \\
%          \midrule
%          \rowcolor{lightgray} \multicolumn{5}{c}{\textbf{IC-DST Baseline}} \\
%        N/A & \SLM & N/A & 43.63 / 87.23 & 74.41 / 82.09 \\
%         \rowcolor{lightgray} \multicolumn{5}{c}{\textbf{Zero-Shot Correction Baseline}} \\
%  x  & \SLM & GPT-4o & 24.35 / 73.32  & 61.28 / 69.52   \\
%         \rowcolor{lightgray} \multicolumn{5}{c}{\textbf{In-Context Correction Baselines}} \\
%        \SLM  & \SLM & \SLM & 43.09 / 87.52  & 72.91 / 83.85   \\
%        \SLM & \SLM & GPT-4o & 58.68 / 93.37 & 85.77 / 91.05 \\
%       \rowcolor{mintgreen} \multicolumn{5}{c}{\textbf{Our Finetuned Correction SLM}} \\
%        \SLM  & \SLM & finetuned Llama & 53.35 / 92.32  & 80.71 / 89.31  \\    \bottomrule
%     \end{tabular}
%     \caption{Results on MultiWOZ 2.4. We use 1\% training set as training data and 10\% testing set as testing data. We report zero-shot correction baseline with GPT-4o and in-context correction baseline with both \SLM and GPT-4o. Our proposed correction model was in-context finetuned with self-generated errors and gold labels.}
%     \label{tab:correctionlm_mwoz}
% \end{table*}

\begin{table*}[t]
    \small
    \centering
    \begin{tabular}{l|l|l|cc}
    \toprule
        \textbf{Inference Model on Train}  & \textbf{Inference Model on Test}  & \textbf{Correction Model}  & \textbf{DST JGA/F1} & \textbf{TLB JGA/F1} \\
         \midrule
         \rowcolor{lightgray} \multicolumn{5}{c}{\textbf{Single Pass ICL Baselines}} \\
       N/A & \SLM & N/A & 40.10 / 87.11 & 72.32 / 81.43 \\
        N/A & GPT-4o & N/A & 47.96 / 90.32 & 81.69 / 87.23 \\
 %        \rowcolor{lightgray} \multicolumn{5}{c}{\textbf{Zero-Shot Correction Baseline}} \\
 % x  & \SLM & GPT-4o & 24.35 / 73.32  & 61.28 / 69.52   \\

        \rowcolor{lightgray} \multicolumn{5}{c}{\textbf{In-Context Correction (Non-Finetuned)}} \\
       \SLM  & \SLM & \SLM & 41.54 / 88.27  & 71.19 / 82.61   \\
       GPT-4o & GPT-4o & GPT-4o & 55.60 / 92.64 & 84.35 / 89.83 \\
        \rowcolor{lightgray} \multicolumn{5}{c}{\textbf{Correction (Finetuned)}} \\
       \SLM & \SLM & Correction \SLM & 52.42 / 91.37 & 81.73 / 88.45 \\
 
      \rowcolor{mintgreen} \multicolumn{5}{c}{\textbf{\correctionlm}} \\
       \SLM  & \SLM & Correction \SLM & 56.20 / 92.97  & 83.44 / 89.71  \\    
       % \LLM  & \LLM & Correction-Tuned Llama & 57.13 / 93.53   & 83.93 / 89.92  \\
        \SLM & \LLM & Correction \SLM & 57.35 / 93.27 & 84.25 / 90.10  \\
       \bottomrule
    \end{tabular}
    \caption{Results on MultiWOZ 2.4 with 5\% training data.}
    \label{tab:correctionlm_mwoz_100p}
\end{table*}

% We report in-context correction baseline with both \SLM and GPT-4o. Our proposed correction model was in-context finetuned with self-generated errors and gold labels.

\begin{table*}[t]
    \small
    \centering
    \begin{tabular}{l|l|l|cc}
    \toprule
        \textbf{Inference Model on Train}  & \textbf{Inference Model on Test}  & \textbf{Correction Model}  & \textbf{DST JGA/F1} & \textbf{TLB JGA/F1} \\
         \midrule
         \rowcolor{lightgray} \multicolumn{5}{c}{\textbf{Single Pass ICL Baseline}} \\
       N/A & \SLM & N/A & 16.55 / 62.80 & 66.57 / 75.53 \\
     N/A & \LLM & N/A & 35.61 / 89.08 & 78.64 / 90.01 \\
 %        \rowcolor{lightgray} \multicolumn{5}{c}{\textbf{Zero-Shot Correction Baseline}} \\
 % x  & \SLM & GPT-4o & 24.35 / 73.32  & 61.28 / 69.52   \\
        \rowcolor{lightgray} \multicolumn{5}{c}{\textbf{In-Context Correction (Non-Finetuned)}} \\
       \SLM  & \SLM & \SLM & 28.07 / 84.99  & 69.94 / 82.52   \\
       %  \rowcolor{lightgray} \multicolumn{5}{c}{\textbf{Correction Baseline (Finetuned)}} \\
       % \SLM & \SLM & Correction-Tuned Llama & 19.82 / 70.02 & xx / xx \\
      \LLM & \LLM & \LLM & 41.18 / 90.65 &  80.54 / 91.86 \\
        \rowcolor{lightgray} \multicolumn{5}{c}{\textbf{Correction (Finetuned)}} \\
       \SLM & \SLM & Correction \SLM & 16.89 / 69.25 & 68.82 / 79.13 \\
 
      \rowcolor{mintgreen} \multicolumn{5}{c}{\textbf{\correctionlm}} \\
       \SLM  & \SLM & Correction \SLM & 37.83 / 87.05  & 79.43 / 89.31 \\    
        \SLM  & \LLM & Correction \SLM & 40.49 / 89.62 & 81.14 / 90.79 \\    
       \bottomrule
    \end{tabular}
    \caption{Results on SGD with 5\% training data. }
    \label{tab:correctionlm_sgd_100p}
\end{table*}

% We report in-context correction baseline with both \SLM and \LLM. Our proposed \correctionlm was in-context finetuned with self-generated errors and gold labels.

\section{Results}

The results for the different approaches are shown in Table~\ref{tab:comparison_mwoz} for MultiWOZ and in Table~\ref{tab:comparison_sgd}.
For both datasets, \correctionlm provides substantial improvements over the single-pass Llama-3-8B baseline. The absolute JGA improvement is 16.1 and 21.3 for MultiWOZ and SGD, respectively, or 40\% and 129\% relative. The more complex ontology leads to lower JGA for SGD due to the large number of null slots, leading to a much larger relative gain for SGD. However, we also see bigger relative gains for SGD than MultiWOZ for TLB JGA  (19\% vs.\ 15\%) and DST F1 (7\% vs. 39\%).

\correctionlm also obtains gains over a single-pass LLM (GPT-4o), particularly for MultiWOZ, with much lower computing costs. For MultiWOZ, \correctionlm outperforms the LLM 2-pass correction approach. For SGD, the 2-pass LLM gives the best results, though \correctionlm is close for TLB.  However, the finetuned SLM gives performance similar to the LLM when correcting the LLM outputs (last row in both tables).

For both datasets, \correctionlm outperforms in-context correction without finetuning, as well as using a finetuned models without examples of corrections.  However, the relative importance of those two aspects of the model differs for the two datasets.  Fine-tuning has a substantial impact for MultiWOZ, but much less for SGD. In contrast, SGD benefits much more from ICL-based correction without finetuning than MultiWOZ. We hypothesize that this difference may be associated with the fact that a large fraction of the SGD test data is from domains unseen in training.
\section{Analysis}

\subsection{Comparisons with SOTAs}
\begin{table*}[t]
    \small
    \centering
    \begin{tabular}{l|l|l|cc}
    \toprule
        \textbf{System}  & \textbf{Backbone LMs}  & \textbf{TeraFLOPs} &  \textbf{DST JGA} & \textbf{TLB JGA} \\
         \midrule
            \rowcolor{lightgray} \multicolumn{5}{c}{\textbf{Previous State-of-the-art}} \\
        IC-DST~\cite{hu2022context} & Codex (175B)  & 22.0 M & 55.43 & N/A \\

        IC-DST 2.0~\cite{hu2022context} & GPT-3.5-Turbo (175B)  & 22.0 M & 49.68 & 78.21 \\
        DS2~\citep{shin2022dialogue} & T5 (220M) & N/A & 49.89  & N/A \\
       % \orchestrallm-SenBERT & T5 (60M), GPT-3.5-Turbo (175B) & 5\% & 8.8M & 50.19 & 80.74 \\
\orchestrallm~\cite{lee2024orchestrallm} & T5 (60M), GPT-3.5-Turbo (175B)  & 8.3 M & 52.68 & 82.46 \\
        RefPyDST~\cite{king2023diverse} & Codex (175B)  &  110.0 M (?) & 
62.30 & N/A \\
      \rowcolor{mintgreen} \multicolumn{5}{c}{\textbf{Ours}} \\
       % \correctionlm  & Llama3 (8B) & 1\% & 1.7M & 51.09  & 80.91  \\    
       \correctionlm  & Llama3 (8B)  & 1.7 M & 56.20  & 83.44  \\    
       \bottomrule
    \end{tabular}
    \caption{Full results with 5\% training data on MWOZ 2.4 comparing \correctionlm to the previous state-of-the-art on low-resource DST: IC-DST, IC-DST 2.0, DS2, \orchestrallm, and RefPyDST. $?$ marks the estimated TeraFLOPs for RefPyDS by assuming prompting a 175B GPT3-style model with 4096 tokens (same as IC-DST and IC-DST 2.0) five times per turn.}
    \label{tab:comparison_mwoz}
\end{table*}
\begin{table*}[t]
    \small
    \centering
    \begin{tabular}{l|l|l|cc}
    \toprule
        \textbf{System}  & \textbf{Backbone LMs} & \textbf{TeraFLOPs} &  \textbf{DST JGA} & \textbf{TLB JGA} \\
         \midrule
        IC-DST 2.0 & GPT-3.5-Turbo (175B) &  121.0 M & 33.15 & 79.47 \\

       \orchestrallm & T5 (220M), GPT-3.5-Turbo (175B) & 52.0 M & 33.07 & 77.12 \\
    \rowcolor{mintgreen} \multicolumn{5}{c}{\textbf{Ours}} \\
\correctionlm  & Llama3 (8B)  & 9.3 M & 37.83  & 79.43 \\    
       \bottomrule
    \end{tabular}
    \caption{Full results with 5\% training data on SGD comparing \correctionlm to the previous state-of-the-art on low-resource DST: IC-DST 2.0, \orchestrallm.}
    \label{tab:comparison_sgd}
\end{table*}
We compare \correctionlm to previous state-of-the-art low resource works. Following prior work~\cite{lee2024orchestrallm}, we evaluate computation efficiency by reporting the aggregate computational cost for performing
inference across the testing dataset, measured in TeraFLOPs. The MultiWOZ results are shown in table~\ref{tab:comparison_mwoz}. RefPyDST~\cite{king2023diverse} is the best system in few-shot settings, but their system requires sampling five candidate dialogue states which significantly increases the computational costs. Furthermore, Codex~\cite{chen2021evaluating} is not an available API model anymore. Compared to \orchestrallm~\cite{lee2024orchestrallm}, we see that \correctionlm can achieve better performance in all metrics and requires much less computation with the trade-off of having higher latency associated with two inference passes. Similar trends for SGD can be observed in Table~\ref{tab:comparison_sgd}.

\subsection{Out-of-Domain Evaluation}
\begin{figure}[t]
\centering
  \includegraphics[width=\linewidth]{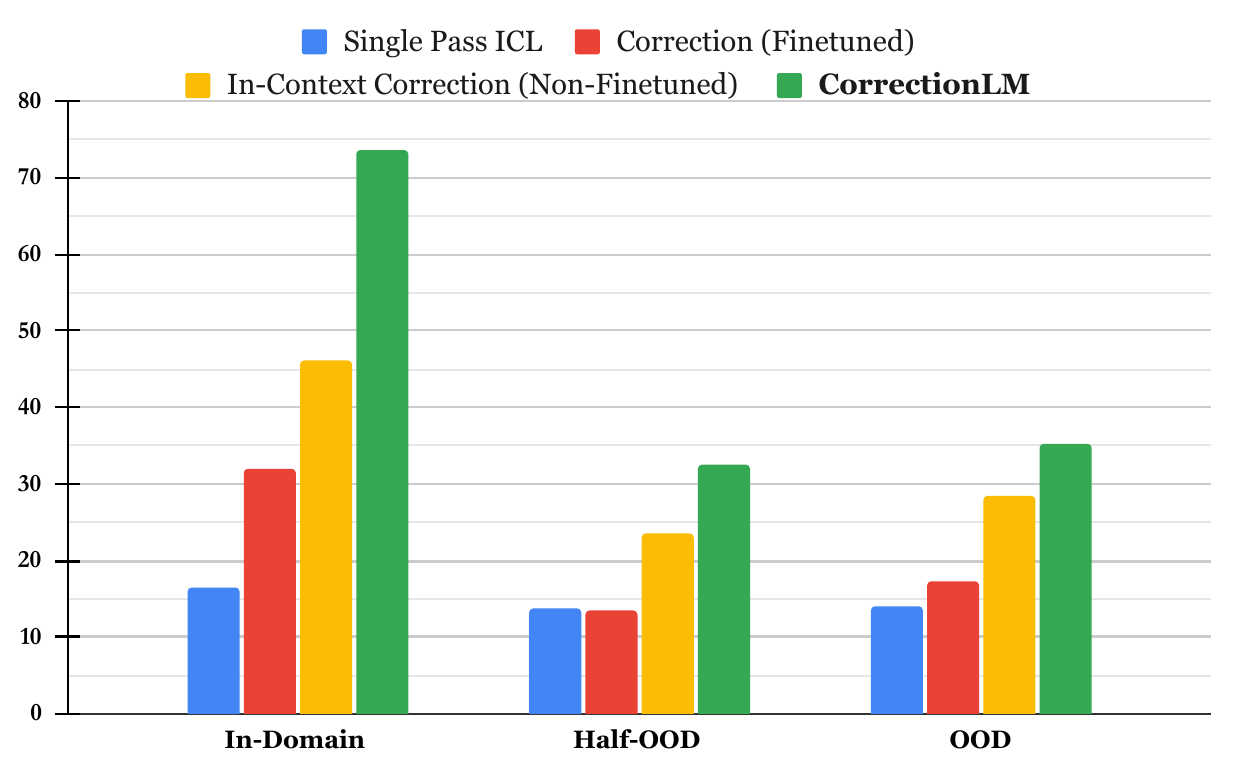}
  \caption{
Cross-domain generalization results on SGD.
We denote In-Domain when all of the testing domains
are in the training set and denote OOD when all of
the testing domains are not in the training set. For all
other dialogues, we categorize them as Half OOD. We
report DST JGA for all settings.} 
  \label{fig:analysis_OOD}
\end{figure}

To evaluate the domain generalization capabilities of \correctionlm and gain deeper insights into the types of errors it corrects, we categorize the SGD dialogues into different error types and present a detailed breakdown in Figure~\ref{fig:analysis_OOD}. There are about 20k out-of-domain (OOD) turns, 18k half-OOD turns and 40k in-domain turns. First, we observe that the correction baselines underperform in OOD scenarios compared to in-domain settings, while \correctionlm demonstrates significantly improved results on OOD corrections. Additionally, \correctionlm far surpasses the non-finetuned in-context correction baseline in in-domain performance, highlighting the effectiveness of our framework in leveraging in-context examples. Finally, the largest performance improvements for \correctionlm are observed in in-domain conversations, further demonstrating its strength in familiar contexts.

\section{Related Work}

\subsection{Correction with Prompting} 
Pioneering studies by \citet{bai2022constitutional} and \citet{ganguli2023capacity} revealed that LLMs have the capacities to self-correct, offering feedback on their own outputs and subsequently refining them based on this self-generated feedback. Building on this, \citet{kim2024language} employed prompting techniques to guide LLMs in generating both feedback and improved outputs, while \citet{madaan2024self} iteratively used in-context exemplars to enhance LLM performance. Additionally, \citet{shinn2024reflexion} explored the correction abilities of LLMs by extending the feedback mechanism to include not just textual feedback but also scalar rewards. In contrast to these approaches, our work does not rely on self-generated or external feedback beyond a limited set of gold-standard answers.

Another line of research has focused on enhancing reasoning frameworks by engaging multiple language model agents in discussions or debates, as seen in studies by \citet{chan2024chateval}, \citet{wang-etal-2024-rethinking-bounds}, and \citet{chen-etal-2024-reconcile}. Our work uses a single SLM to achieve self-correction, without the need for complex multi-agent interactions.

\subsection{Correction with External Tools}
Correction models often integrate information from external tools, such as code executors \citep{chencodet,chen2024teaching}, mathematical proof assistants \citep{first2023baldur}, and search engines \citep{gao2023rarr,gou2024critic}. Unlike these approaches, our work is self-contained, relying solely on a small set of gold-standard answers without the assistance of external tools.

\subsection{Correction with Finetuning} 
Finetuning strategies have been extensively explored to enhance correction performance beyond what prompting alone can achieve. \citet{saunders2022self} and \citet{huang2023large} have finetuned LLMs to correct their own outputs, though this process is computationally intensive. To mitigate these challenges, recent studies have focused on finetuning SLMs for self-critique and self-correction \citep{shridhar-etal-2023-distilling,selfee2023,yu-etal-2024-teaching}. While \citet{shridhar-etal-2023-distilling} and \citet{selfee2023} depend on LLM-generated outputs, including initial predictions, feedback, and refined results, \citet{yu-etal-2024-teaching} address training-test misalignment by leveraging SLM-generated initial predictions. Similar to our work, \citet{welleck2023generating} train correction models using a few selected examples, but their approach relies on external evaluators for example selection. In contrast, our work uses semantic embeddings for example retrieval, eliminating the need for external evaluation.

Further differentiating our approach, \citet{paul2024refiner} generate feedback data from GPT-3.5 by synthesizing human-defined error types and heuristics, whereas our work synthesizes data using an in-context learning framework with SLMs, bypassing the need for human-authored rules. Finally, \citet{akyurek-etal-2023-rl4f} employ reinforcement learning to train critique models that maximize GPT-3's end-task performance. Our method, however, focuses on supervised fine-tuning for effective and efficient self-correction.

\subsection{Language Models for Dialogue State Tracking}
The earliest systems finetuned from autoregressive pretrained LMs were developed to eliminate the need of task-specific modules or classifiers~\cite{ham2020end,hosseini2020simple,peng2020soloist}. Subsequent to the publication of ~\cite{lee2021dialogue}, other systems that are finetuned from sequence-to-sequence pretrained LMs have been developed to build~\cite{bang2023task,imrattanatrai2023end,su2022multi}. 

To reduce the need for labeled data in DST, few-shot methods based on fine-tuning pretrained LMs have been proposed~\citep{wu2020improving, li2021zero, su2022multi, shin2022dialogue, lin2021leveraging, xie2022unifiedskg}. However, these systems are less flexible as they require retraining when new slots or domains are added, and fine-tuning LMs is computationally expensive. To address these challenges, ~\citet{xie2022unifiedskg} and ~\citet{madotto2021few} were the first to apply ICL for few-shot DST, but their systems underperformed compared to previous fine-tuning-based methods. ~\citet{hu2022context}, is the first successful system to apply ICL to DST. ~\citet{king2023diverse} further improved this approach by reformulating ICL as a Python programming task and diversifying the retrieved in-context exemplars. ~\citet{lee2024orchestrallm} also develop a routing strategy to simultaneously leverage a small language model and a large language model to achieve strong few-shot performance.

\section{Conclusion}
In this paper, we propose \correctionlm, a self-correction framework that efficiently finetunes a SLM with correction-augmented in-context exemplars. Experimental results show that \correctionlm can make successful refinements over a strong LLM-based in-context learning baseline, with significant improvements at both TLB and DST levels, and at a substantial savings in computation. Using multiple metrics, we find larger gains for the more complex SGD task. Both finetuning and the use of in-context examples benefit the model, but finetuning has the most benefit for in-domain data.
%40\% relative DST JGA for MultiWOZ and over 120\% relative DST JGA for SGD. When compared to the baseline SLM, the improvements are much greater, particularly for the more challenging SGD data set.
\section{Limitations}
A potential limitation of \correctionlm is its reliance on the quality of in-context exemplars, which may limit performance if examples are not adequately representative or relevant for correction tasks. Additionally, its generalizability to complex tasks like coding or mathematical reasoning remains untested, as our evaluations are limited to two DST benchmarks.

% Entries for the entire Anthology, followed by custom entries
\bibliography{anthology,custom}
\pagebreak
%\onecolumn
% \appendix
% \input{sections/appendix}

\end{document}